\documentclass[10pt, journal, compsoc]{IEEEtran}
\IEEEoverridecommandlockouts

\usepackage{enumitem}
\usepackage{cite}
\usepackage{amsmath}
\usepackage{amsmath,amssymb,amsfonts}
\usepackage{alphabeta}
\usepackage{algorithmicx}
\usepackage{algorithm} 
\usepackage{algpseudocode} 
\usepackage{float}
\usepackage{graphicx}
\usepackage{textcomp}
\usepackage{xcolor}
\usepackage{threeparttable}
\usepackage{multirow}
\usepackage{tabularx}
\def\BibTeX{{\rm B\kern-.05em{\sc i\kern-.025em b}\kern-.08em
    T\kern-.1667em\lower.7ex\hbox{E}\kern-.125emX}}

\begin{document}
\title{Determining Intent of Changes to Ascertain Fake Crowdsourced Image Services}

\author{Muhammad~Umair,
        Athman~Bouguettaya,~\IEEEmembership{Fellow,~IEEE}
        and~Abdallah~Lakhdari
\IEEEcompsocitemizethanks{\IEEEcompsocthanksitem M. Umair, A. Bouguettaya, A. Lakhdari are with the School of Computer Science,
University of Sydney, Australia.
E-mail: \{muhammad.umair, athman.bouguettaya, abdallah.lakhdari\}@sydney.edu.au}
}

 
\IEEEtitleabstractindextext{%
\begin{abstract}

We propose a novel framework for crowdsourced images to determine the likelihood of an image being \textit{fake}. We use a service-oriented approach to model and represent crowdsourced images uploaded on social media, as \textit{image services}. Trust may, in some circumstances, be determined by using only the non-functional attributes of an image service, i.e., image metadata. We define \textit{intention of changes} as a key parameter to ascertain fake image services. A novel framework is proposed to estimate the intention of underlying changes considering change in semantics of an image. 
Our experiments show high accuracy using a large real dataset.
\end{abstract}
}

\maketitle

\section{Introduction}
Social media has become a key platform to share news and information related to real-life events. There are approximately 5 billion active users on social media~\cite{griffis2014use}. Social media users publish a large amount of data related to public events~\cite{liu2011using}. These crowdsourced social media streams may possess some critical information that describes a situation from different perspectives~\cite{rosi2011social}. For instance, a social media stream may unveil some unfolding situations in a road accident~\cite{gu2016twitter}. Utilizing these crowdsourced images can significantly facilitate the task of scene analysis and reconstruction.


We use the term \textit{image service} to refer to a social media image. In this respect, we define an image service to have \textit{functional} and \textit{non-functional} properties. 
Functional attributes are related to the capturing of an image, e.g., switching picture/video modes, panorama mode, delayed/timed picture taking, etc.
Whereas, non-functional attributes help in distinguishing different image services, e.g., subject distance, camera elevation angle, resolution, color space, etc.

Existing work on image services focuses on image service selection and composition techniques for scene analysis and reconstruction~\cite{aamir2018social}. 
A fundamental assumption in the existing scene reconstruction approaches has so far been that image services are intrinsically \textit{trustworthy}. However, the \textit{trust} issue is paramount when image services are assumed to be crowdsourced~\cite{shen2019fake}. 
Defining trust in crowdsourced environments has traditionally been addressed using natural language processing and information retrieval techniques~\cite{gupta2013faking}. These approaches are usually costly in terms of computation and resources.
A preliminary work on a service-based trust framework for image services is described in~\cite{aamir2018stance, aamir2018trust}. This particular work is based on the users' comments in an image service to assess its credibility~\cite{aamir2018trust}. 
The credibility of image services may not be completely assessed based only on the user's stance~\cite{colliander2019fake}. 

We propose to assess trust using an objective framework consisting of modifications and updates in an image service. Changes in an image service are often reflected in its non-functional attributes~\cite{ghosh2017detection}. For instance, a falsely claimed location in a picture can be confirmed using the GPS and other spatial tags in the image metadata~\cite{riggs2018image}. We leverage only the non-functional attributes to determine the likelihood of an image to be \textit{fake}. \textit{We define fake as a context-sensitive parameter that reflects whether the changes in an image are harmful for a particular application.} For instance, an image which is considered as fake for reconstructing a road accident scene may not be fake for analyzing a public march. Therefore, changes in image services should be extensively investigated before concluding on the fakeness of an image. In this respect, we propose a detailed framework to analyze modifications in image services. 
We approach this problem from social media owners perspective implying that we have access to the metadata and ground truth of the image.

We define \textit{intent of changes} as a key parameter to ascertain fake image services. Intention reflects the purpose of introducing changes in images. Changes can be either ill-intentioned or well-intentioned. For instance, hiding the facts in an image service are usually ill-intentioned changes whereas improving the outlook of an image is mostly considered as a well-intentioned change. We propose to estimate the intention of modifications by determining the change in semantics of an image. In this respect, we introduce a novel way of employing Latent Semantic Analysis (LSA) to extract semantics of an image service. LSA forms image vectors with the values representing the importance of different latent contexts in an image. Cosine similarity measure is then applied on these image vectors to quantify semantic differences among images. We formalize intention in terms of these semantic changes in an image. Afterwards, the intention is used to determine the likelihood of an image being fake in a specific context. The proposed method is valid for a typical set of modifications that are reflected in non-functional part of an image service. Some changes may not be reflected in non-functional part of an image service, e.g., changes in shades and intensity of the colors. We consider only those changes which are reflected in the non-functional attributes specifically the spatio-temporal and contextual attributes. 
Experiments are conducted on two different large real datasets for evaluation. \color{black}We use accuracy and time efficiency metrics to evaluate the proposed framework. It achieves 80-95\% accuracy on a systematic set of experiments. The following are our main contributions: 

\begin{itemize}
    \item A novel services-based framework is proposed to determine fake image services. The framework only relies on the non-functional attributes of an image service and does not require processing the image.
    \item Intent of changes is defined as an imperative parameter to ascertain fake image services. A novel way is introduced to formalize intention based on the change in semantics of an image service.
    \item LSA is employed in a unique way to determine semantics of an image and to quantify the semantic changes.
\end{itemize}
\color{black}


\section{Motivating Scenario}

\begin{figure*}[!h]
\centerline{\includegraphics[width = \textwidth]{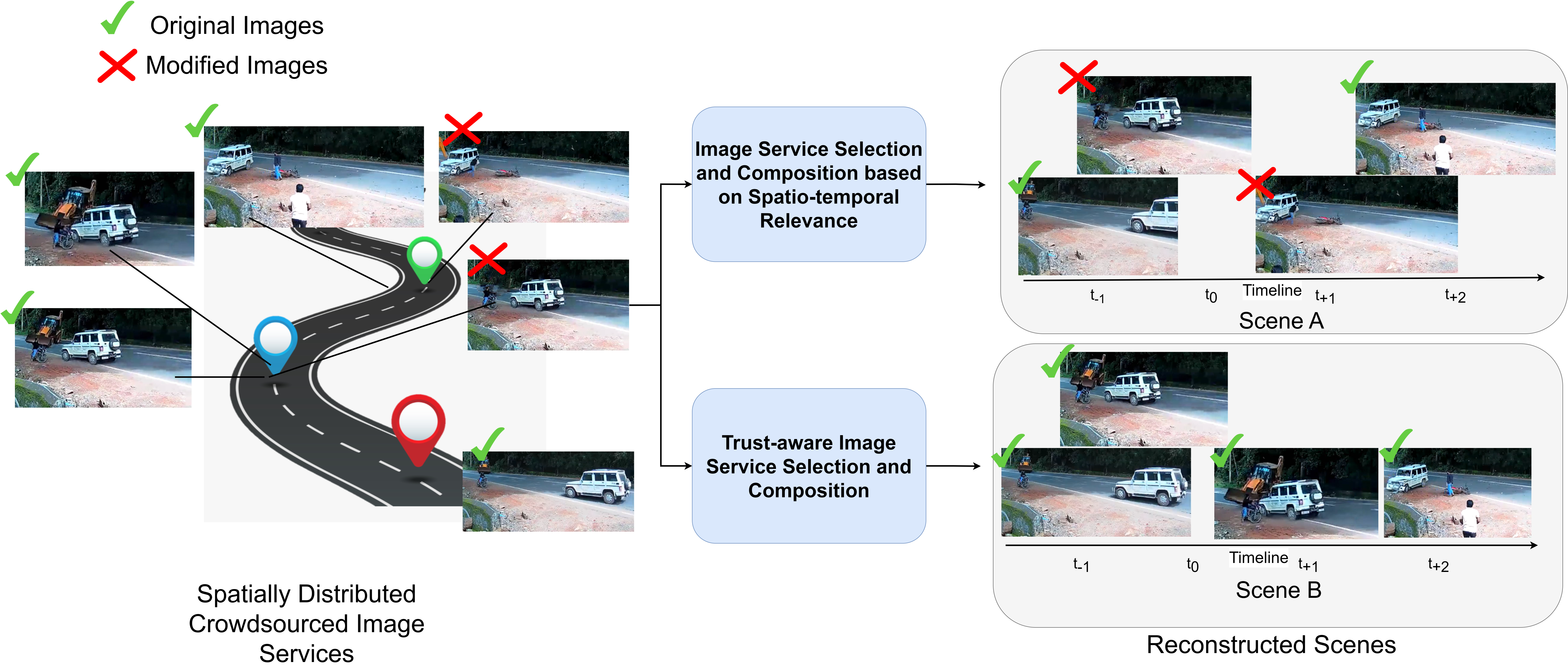}}
\caption{Motivating Scenario}
\label{motivation}
\end{figure*}

We use a scene of a road accident as our motivating scenario. Figure~\ref{motivation} shows images of a road accident uploaded on social media. Before the accident, a bike rider was standing at a side of a road while a bulldozer was working on the other side. The bulldozer lost control and started moving towards the bike rider. The bike rider was about to be hit by the bulldozer while a speeding Sport Utility Vehicle (SUV) crashed into the bulldozer right before him (see original images in Figure~\ref{motivation} marked with the green ticks). We take this scene as an example to demonstrate how the modifications may lead to erroneous conclusions. In this respect, we remove the bulldozer from some images (modified images in Figure~\ref{motivation}). Modified images give an impression that the accident took place due to the negligence of the SUV's driver. 

We apply the existing service-oriented approach proposed in \cite{aamir2020social} to compose relevant image services. Non-functional attributes of image services are only utilized in these methods to reconstruct the scene. These scene reconstruction techniques are based on the spatio-temporal relevancy of the image services. They do not consider modifications in image services and hence, are unable to filter out the modified images. Scene A in Figure~\ref{motivation} is reconstructed by the existing image service selection approaches. 
It can be observed that the scene consists of some modified images giving a wrong impression that the accident took place due to the negligence of the SUV's driver. 

There is a need to intrinsically evaluate the image services before using them for scene reconstruction. 
The proposed framework 
can be utilized to filter out the untrustworthy images while reconstructing a scene (see Figure~\ref{motivation}). The scene reconstructed using the proposed framework only consists of trustworthy images (Scene B). Scene B shows that a bulldozer is also involved in the accident and the accident was not entirely due to the negligence of Bolero's driver.

\section{Related Work}
Scene analysis and reconstruction approaches are mostly based on image processing and object identification~\cite{chae2012spatiotemporal}. These solutions are usually costly and computationally intensive~\cite{slabaugh2001survey}. Whereas, existing work on image services is related to image service selection and composition strategies. 
A novel social-sensor cloud service model is proposed in~\cite{aamir2017social2}. The proposed approach conceptualizes the spatio-temporal and textual aspects of social-sensor data streams as image services’ functional attributes, and the qualitative aspects as their non-functional attributes. The proposed framework is a 4-stage algorithm capable of context-aware selection of social-sensor cloud services. It composes image services to form a mosaic/tapestry in the spatial aspect and a storyboard in the temporal aspect. An image services composition model is proposed in~\cite{aamir2018social}. The proposed model composes image services based on user queries. The model relies only on the meta-data and related posted information with the image. A heuristics-based social-sensor cloud service selection and composition model is proposed in~\cite{aamir2020social}. 
It proposes a context and direction aware spatio-temporal clustering and recommendation approach. It helps to compose the relevant image services to form a tapestry in the spatial aspect and a storyboard in the temporal aspect. A fundamental issue in these frameworks is that they do not consider trustworthiness of crowdsourced image services. 

Image services in a crowdsourced environment can be untrustworthy. Traditional approaches to identify untrustworthy images are based on image processing and machine learning~\cite{patel2022fake}. A multi-modal approach is proposed in \cite{singh2021predicting} that utilizes new and upgraded models to detect fake images shared over social media platforms. A deep learning-based approach for detecting fake images is proposed in~\cite{hsu2020deep}. Neural networks and image processing is used in~\cite{singh2021image} to spot fake images shared over social media platforms. Traditional digital forensics and artificial intelligence is used in~\cite{yang2021detecting} to identify forgery in images. A block-based approach for Copy-Move Forgery detection is implemented in~\cite{kour2022fast}. It uses quantized discrete cosine transform coefficients to detect forgery over an image. Histogram-based fake colorized image detection and feature encoding-based fake colorized image detection is proposed in~\cite{guo2018fake}. Robust hashing based fake image detection is proposed in\cite{tanaka2021fake}. 
The aforementioned image processing based techniques have high accuracy in determining the fakes in an image but require high computational power and memory.

Some recent studies claim that a subset of \textit{trust} can be derived using light weight service-oriented approaches~\cite{aamir2018stance, aamir2018trust, umair2023detecting}. A new image services \textit{trust} model is proposed in~\cite{aamir2018trust}. The trustworthiness of an image service is measured based on the users' stance. Textual features of the image services, i.e., comments and meta-data, e.g., spatio-temporal information are utilized to gather the trust-rate of the service. A users’ stance and credibility based image service's trust model is proposed in~\cite{aamir2018stance}. The proposed model considers various indicators such as the stance embedded in the services’ comments, their meta-data, e.g., time, along with the users’ credibility. It models the interactions between commenters and sub-comments in terms of their comments. Natural Language Processing is leveraged in such solutions to classify text in a comment~\cite{umair2020multi}. These approaches are unable to capture modifications in an image service because the misleading content on social media may receive positive comments from other users~\cite{colliander2019fake}. Moreover, comments from credible users can also be biased~\cite{zimmer2019fake}. Therefore, we propose a more objective trust model for image services which also considers modifications and updates in an image to determine its credibility. The proposed framework is unique in a sense that it does not require the actual images to be processed to determine their trustworthiness.

\section{Image Service Model}\label{sec:imgsrvModel}


We abstract an \textit{image as a service} having functional and non-functional attributes. 
Functional attributes represent the actions involved in capturing an image. 
\begin{table*}[!h]
\caption{Description of Metadata Tags}
\begin{tabular*}{\textwidth}{@{} p{2.6cm}p{4cm}p{3cm}p{4cm}l @{} }
\hline
\textbf{Categories} & \textbf{Description} & \textbf{Possible Modifications} & \textbf{Impact} & \textbf{List of Attributes} \\ \hline

\multirow{2}{4cm}{Spatial Features} 
& \multirow{2}{4cm}{Spatial metadata tags describe the location at which the image was taken.} & \multirow{2}{3cm}{Incorrect location, Fake background} & \multirow{2}{4cm}{Forged location tags lead to the selection of spatially incorrect images.} & GPS Longitude \\  
  &&&&GPS Latitude \\ &&&&GPS Satellite \\ &&&&City, Country \\  &&&&Location \\ &&&&State  \\  \\ 
 
\multirow{3}{4cm}{Temporal Features}
& \multirow{2}{4cm}{Temporal metadata tags describe the date and time when the image was taken.} & \multirow{2}{3cm}{Outdated images, Manipulations in the objects that reflect time i.e., clock, sun, and moon etc.} & \multirow{2}{4cm}{Tampered timestamps lead to the development of a fake story.} & Date Digitized \\  
  &&&&GPS Timezone Offset \\ &&&&GPS Timestamp\\ &&&&GPS Datestamp \\ &&&&Date Time Original  \\ \\  
  
\multirow{4}{4.2cm}{Contextual Features}
& \multirow{2}{4cm}{Contextual features define the context of an image. Context of a picture is usually described by textual information available with the image. However, contextual attributes may also contain some other metadata tags i.e., details of the ambience and surroundings.} & \multirow{2}{3cm}{Misleading perception consistent with modified objects and modified background} & \multirow{2}{4cm}{Fake context may support the existence of fake objects in an image.} & Title \\  
  &&&&Caption \\ &&&&Content Description\\ &&&&Contribution Description \\ &&&&Headline \\ &&&&Image Description \\ &&&&Keywords \\ &&&&Instructions \\ &&&&Weather Profile \\ &&&&Semantic Names \\  \\  

 
 \multirow{2}{4cm}{Image Quality Features} 
& \multirow{2}{4cm}{These metadata tags are related to the quality of an image.} & \multirow{1}{3cm}{Cropped Picture, Blurred objects} & \multirow{2}{4cm}{Image quality attributes are being modified to either enhance the picture or hide the facts.} & Resolution \\  &&&&Coverage \\
     &&&&White Balance\\  &&&&Subject Distance  \\ &&&&Camera Elevation Angle \\ \\

\hline
 
\end{tabular*}
\label{FeatureVariables}
\end{table*}
Whereas, non-functional part of an image service can be represented as a function of spatio-temporal and contextual attributes: 
\begin{equation}
    nf = \{\zeta, \tau, c, \iota\} 
    \label{eq2}
\end{equation}
where \textit{$ζ$} represents the set of spatial attributes, \textit{$τ$} is the set of temporal attributes, \textit{c} contains contextual attributes and $ι$ represents intrinsic attributes. 


\subsection{Non-functional Attributes of an Image Service}

We identify different non-functional attributes of an image service that may indicate changes within an image (see Table~\ref{FeatureVariables}). These \textit{metadata} tags are categorized based on the information they contain. This information, if modified, is an indication of modifications within the image. It is worth mentioning that some social media platforms remove public access to this metadata. The proposed framework approach this problem from social media owners perspective that allows us to assume that we have access to the metadata. The proposed framework is therefore intended to serve the social media owners.
We group the non-functional attributes into the following two main categories:
\subsubsection{Extrinsic Attributes}
\textit{Extrinsic non-functional attributes} usually reflect modifications which are not within an image i.e., date, time and location etc. Extrinsic non-functional attributes of an image service can be grouped into the following subcategories:

\begin{figure*}[!h]
\centerline{\includegraphics[width = 0.7\textwidth]{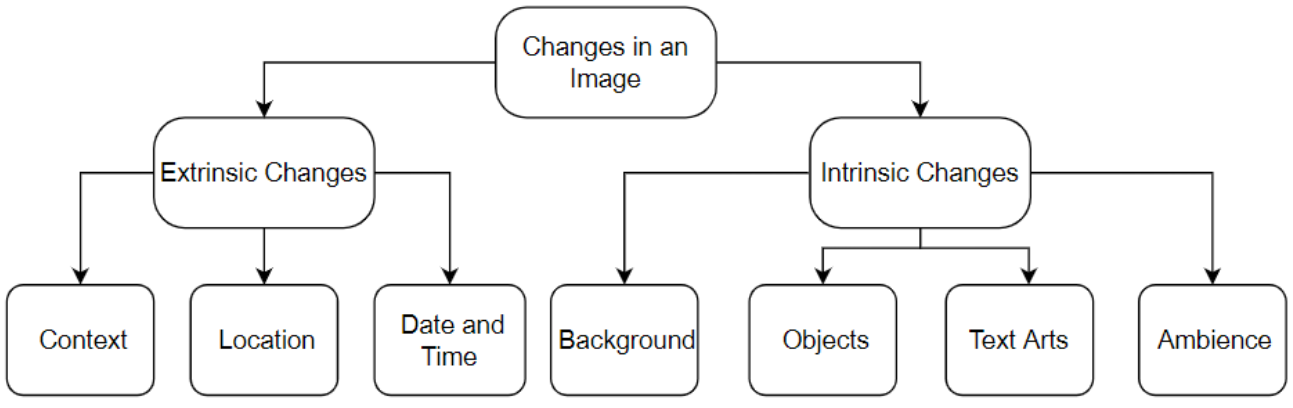}}
\caption{Categorization of Potential Modifications in an Image}
\label{fig1}
\end{figure*}

\begin{itemize}

    \item \textit{Spatial Features}: Spatial attributes represent the location where the image was captured. Location tags help in the selection of spatially relevant images for a scene reconstruction. Modified spatial tags may be an indication of fake background in the image~\cite{elmaci2021comparative}. For instance, a footage of a car accident can be forged to display a fake background. It can make the audience perceive that the car was at a different location while the accident took place. Changes in the spatial attributes are denoted by $Δζ$.

    \item \textit{Temporal Features}: Temporal attributes represent the date and time when the image was captured. Date and time have no direct relation with the credibility of an image. However, it may affect the reconstructed scene because of the modified timeline~\cite{padilha2021content}. Temporal metadata tags are being forged to develop a fake story line. For instance, in case of a car accident, timestamp of the captured scene can be tampered to depict a wrong cause of the accident. Changes in temporal attributes are represented by $Δτ$.
    
    \item \textit{Contextual Features}: Contextual features are related to the context of an image. It is often available in form of text or metadata tags. Contextual features help in the selection and composition of relevant image services while reconstructing a scene~\cite{aamir2018social}. Fake textual description may support the fake temporal and spatial tags of an image~\cite{shen2019fake}. For instance, a scene of a car accident can be described in a way consistent with the fake temporal and spatial tags. Contextual features may also contain description about the surroundings. Contextual changes are represented by $Δc$.
   
\end{itemize}

\subsubsection{Intrinsic Attributes}
\textit{Intrinsic non-functional attributes} usually reflect modifications introduced within an image. For instance, \textit{shutter speed} and \textit{exposure time} are two intrinsic attributes which reflect the proportion of light in an image. These two attributes are dependent on each other. Higher the shutter speed, the smaller will be the exposure time~\cite{simon2022automatic}. Any discrepancy among these two attributes may indicate some lighting touch-ups in an image. 
We consider changes in spatial, temporal, contextual and intrinsic attributes to formalize modifications in an \textit{image service}. Equation~\ref{eq3} can be used to represent changes in an image service as follows:

\begin{equation}
    \Delta ImgServ = \{\Delta \zeta, \Delta \tau , \Delta c, \Delta \iota\} 
    \label{eq3}
\end{equation}

\subsection{Categorization of Modifications in Image Services}\label{Modifications}

We consider all those changes in an image that are reflected in one or more non-functional attributes. 
In this respect, we propose a categorization that 
is based on the nature of changes in the non-functional attributes.
The proposed categorization distinguishes different types of changes that may require a different treatment to determine the severity of changes. We categorize the possible modifications in an image into different groups as shown in Figure~\ref{fig1}. These changes in an image service are described as follows:

\begin{figure}[!h]
\centerline{\includegraphics[width = 0.9\columnwidth]{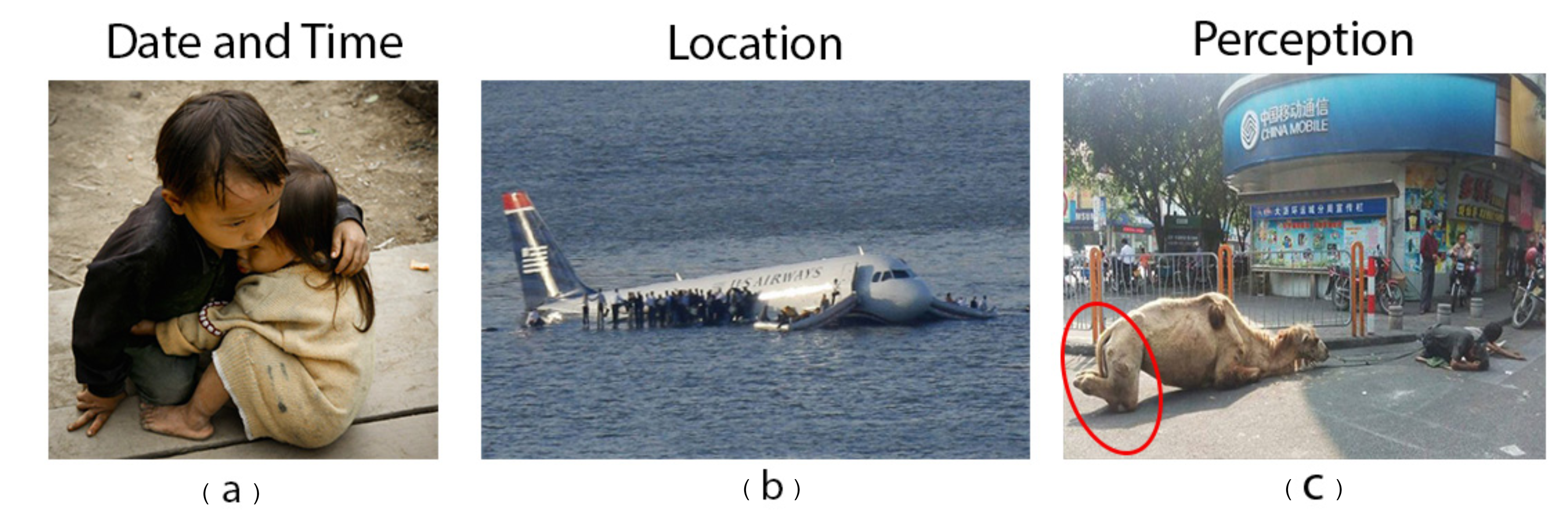}}
\caption{Possible Extrinsic Changes in Images}
\label{fig2}
\end{figure}

\begin{figure}[!h]
\centerline{\includegraphics[width = 0.8\columnwidth]{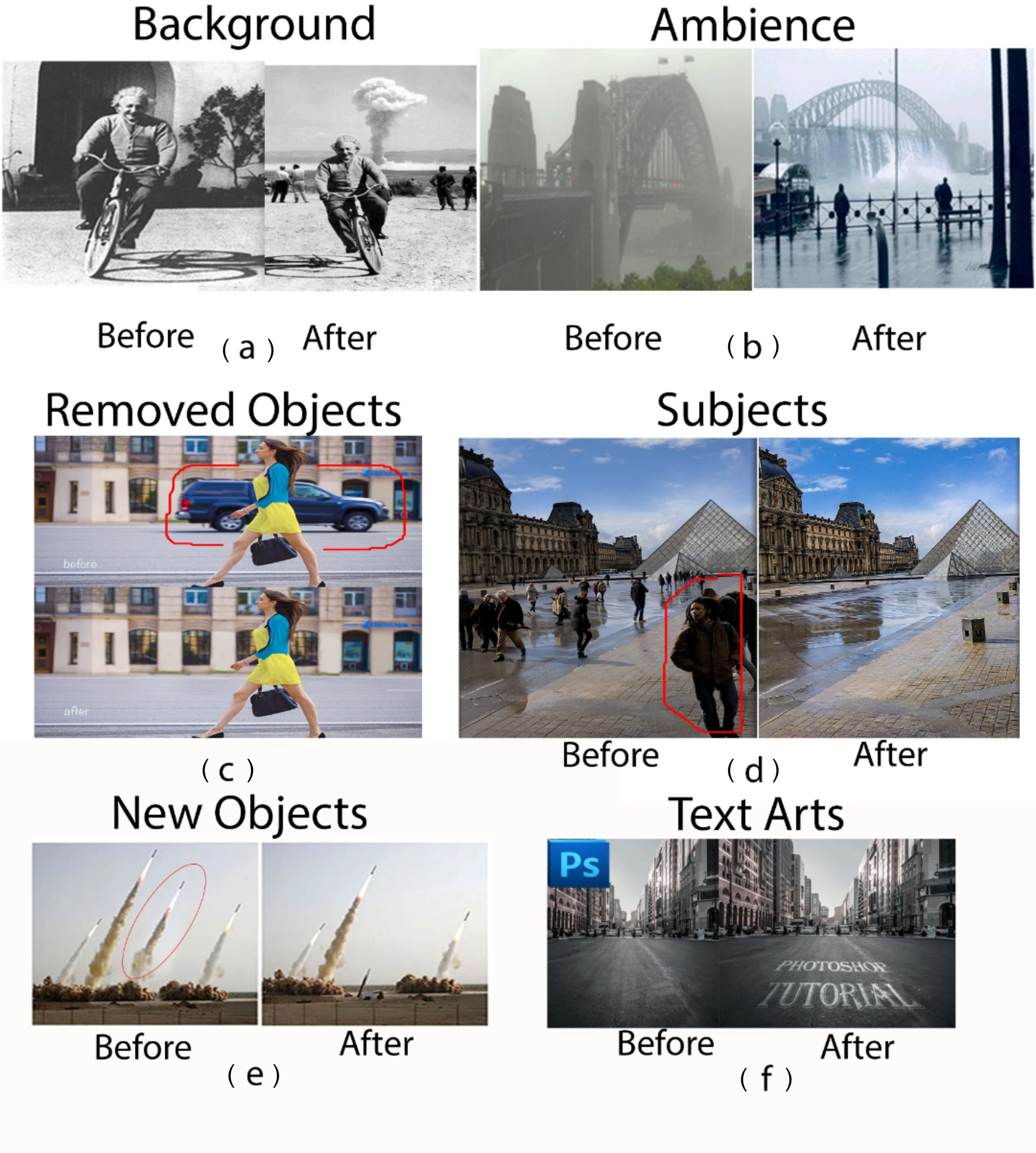}}\caption{Possible Intrinsic Changes in Images}
\label{WellIntInt}
\end{figure}

\subsubsection{Extrinsic Changes}\label{ExtrinsicChanges}
We define \textit{extrinsic changes} as modifications in an image's extrinsic non-functional attributes. Extrinsic changes are not within an image, e.g, changes in location, timestamp, textual description, etc. The image itself remains unchanged. 

\begin{itemize}
    \item \textit{Date and Time}: Temporal attributes of crowdsourced image services may be modified to depict a different temporal information. It has been observed in recent years that outdated images are being shared for a recent event. Figure~\ref{fig2}a claimed to be two children in 2015 Nepal earthquake. It is actually a picture of two Vietnamese taken in 2007. 
    The image contains modified temporal and spatial tags and may mislead the viewer.

    \item \textit{Location}: An image can be doctored extrinsically to represent inaccurate location. The image itself remains unchanged. For instance, Figure~\ref{fig2}b was a viral photo in 2014 claiming the picture of the lost Malaysian MH370 plane. It turned out to be a photo of a plane crash in New York in 2009.
    
    \item \textit{Context}: Contextual changes may change the perception about a picture. Figure~\ref{fig2}c claims a camel with limbs cut off used for begging. The camel is actually resting with legs bent under itself. 
    The picture is unmodified but described in a way to support the misleading perception.
    
\end{itemize}

\subsubsection{Intrinsic Changes}
We define \textit{intrinsic changes} as modifications in intrinsic non-functional attributes reflecting changes within an image. 
The following types of intrinsic changes may exist in an image:
\begin{itemize}
    \item \textit{Background}: Background of an image can be replaced. Spatial and other non-functional attributes may reflect changes in the background. Non-functional attributes may also be modified to make them consistent with the modified background. Inconsistent spatial attributes may be an indication of a modified background. Similarly, inconsistencies in image's quality attributes may also reflect changes in the background. Figure~\ref{WellIntInt}a shows an example photograph with tampered background.

    \item \textit{Ambience}: Introducing clouds, rain, and snow etc., makes an image attractive. Editing the ambience with good intentions is not supposed to tamper the relevant metadata tags. These changes are often reflected in the contextual attributes. Image's non-functional part can be edited to make it consistent with the modified ambience of an image. There are many examples of viral images depicting manipulated disastrous weather. These images may create panic and unrest among people. 
    Figure~\ref{WellIntInt}b is a tailored image showing an exaggerated scene of Sydney storm in 2018. 

    \item \textit{Objects in a Picture}: Photo editors tend to remove unnecessary objects from a picture to improve the background. Figure~\ref{WellIntInt}c shows a car being removed from a photo. Similarly, people have been removed from a scene in Figure~\ref{WellIntInt}d. Removal of objects may be reflected in different intrinsic non-functional attributes. The removed objects may contain critical information. Removing these objects may make these pictures impractical for reconstructing a specific scene. 
    An image may also contain fake objects. Image's non-functional part can be modified to make them consistent with the fake objects in an image. There can be many reasons for the people to add fake objects in an image. For instance, Figure~\ref{WellIntInt}e shows an Iran’s display of military might in 2008. It was doctored to remove a launcher which failed to fire and replaced with a fourth projectile. 
    
    \item \textit{Text Arts}: Text Arts are widely used while editing a picture. Figure~\ref{WellIntInt}f shows a text placed on a road. 

\end{itemize}

\section{Proposed Framework}

\begin{figure*}[htbp]
\centerline{\includegraphics[width = 1\textwidth]{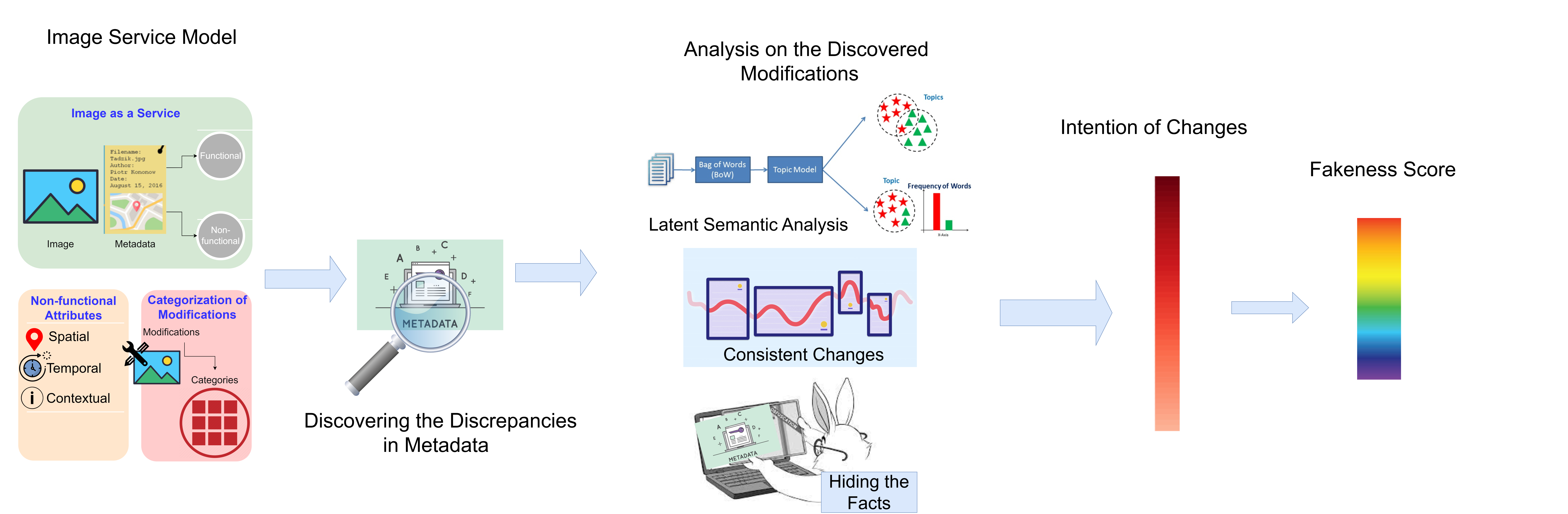}}
\caption{Proposed System Model}
\label{fig:framework}
\end{figure*}

A novel framework is proposed in this section to determine the likelihood of an image to be fake. 
Figure~\ref{fig:framework} is a visual demonstration of the proposed framework.

\subsection{Computing Semantic Changes in Image Services}
 
This section presents mathematical formalization of image service modifications. We primarily focus on the extrinsic modifications. The formalization of intrinsic changes is not covered in this paper. We formalize the modifications in different attributes of an image service into mathematical expressions. These expressions are utilized afterwards to determine the intention of underlying modifications.

Changes in image services can be both semantic (qualitative) and quantitative in nature. For instance, the changes in an image caption is a semantic change whereas temporal changes are usually quantitative. In this section, we represent semantic changes with `δ' and discrete changes with `Δ'. In rest of the sections, `Δ' refers to overall change (both semantic and quantitative) in an attribute.

\subsubsection{Semantic Differences}
Changes in an image service may change the semantics of an image. To compute semantic change in an image, we first need to extract the context of an image and then quantify the change in semantics. We employ LSA in a novel way that it forms image vectors with the context of an image embedded in the vector values. 
In this respect, the attributes are first converted in to image-attribute matrix ($M_{m\times n}$, where m represents number of images and n represents number of attributes). We apply standard Single Value Decomposition (SVD) on image-attribute matrix to find latent topics/contexts in images. It also reduces the number of columns from \textit{m} to \textit{r} (number of latent concepts).

\begin{equation}
    M_{m\times n} = U_{m\times r}Σ_{r\times r}V^t_{r\times n}
    \label{eq10}
\end{equation}

where U is an \textit{$m\times r$} image-aspect matrix with rows representing images and columns representing latent topics/concepts, $Σ$ is an attribute co-occurrence matrix with non-negative real numbers on the diagonal, V is an \textit{$r\times n$} attribute-aspect matrix with rows
representing attributes and columns representing latent topics/concepts, and \textit{r} represents the number of latent concepts/topics. 

Matrix U contains image vectors with the weights of latent topics/concepts embedded in the vector values. SVD can be applied on spatio-temporal attributes separately to get two image-aspect matrices. We use the cosine similarity to quantify the differences in spatio-temporal attributes:

\begin{equation}
    \delta \overrightarrow{t} = 1 - \frac{\sum_{1}^{r} \overrightarrow{t_o} \overrightarrow{t_m}}{\sqrt{\sum_{1}^{r} \overrightarrow{t_o^2}} \sqrt{\sum_{1}^{r} \overrightarrow{t_m^2}}}
    \label{eq11}
\end{equation}

where t\textsubscript{o} represents the original temporal attribute, t\textsubscript{m} represents the modified temporal attribute and \textit{r} is the number of latent topics. Cosine Similarity can be applied on the spatial attributes in the same way as follows:

\begin{equation}
    \delta \overrightarrow{S} = 1 - \frac{\sum_{1}^{r} \overrightarrow{S_o} \overrightarrow{S_m}}{\sqrt{\sum_{1}^{r} \overrightarrow{S_o^2}} \sqrt{\sum_{1}^{r} \overrightarrow{S_m^2}}}
    \label{spatialDist}
\end{equation}

where S\textsubscript{o} represents the original spatial attribute, S\textsubscript{m} represents the modified spatial attribute and \textit{r} is the number of latent topics. Same notations apply to contextual attributes.

\subsubsection{Quantitative Differences}
Most of the image service changes are qualitative in nature. However, in some circumstances where minor changes in spatio-temporal attributes are introduced to compromise the accuracy, we may need to find the quantitative difference between attributes. For instance, minor quantitative temporal differences in a crime scene may impact the evidences. We employ the Manhattan distance metric to assess disparities among discrete temporal attributes. Calculating Manhattan distance is computationally less expensive compared to other distance metrics like Euclidean distance which involves square roots. Previous studies have demonstrated the efficacy of employing simple anomaly detectors based on the Manhattan distance metric, particularly in the context of image data~\cite{li2020low}. Another reason of using Manhattan distance is that the discrete temporal data can be in different units. Moreover, Manhattan distance considers the difference in each component (year, month, day, hour, etc.) separately. This aligns with our natural understanding of time where each component progresses linearly. Another rationale behind using Manhattan distance is its low sensitivity to extreme values in one component. For example, a large difference in years won't heavily outweigh a small difference in minutes. The following equations shows how we calculate Manhattan distance:

\begin{equation}
    \Delta t = \sum_{i=1}^{n} |t_o [i] - t_m [i]|
    \label{eq9}
\end{equation}
where \textit{n} represents the number of discrete temporal attributes. It is worth mentioning that although a pronounced jerk in Manhattan space represents a high magnitude change but it does not always mean a severe change. Severity of changes depends on change in semantics of an image.

Spatial attributes can also be modified to compromise on the accuracy. 
We use haversine formula to determine the distance between two discrete spatial attributes because it results in the shortest distance between two points on a sphere:

\color{black}
\begin{equation}
    a = \sin^2(\Delta \phi/2) + cos\phi_{1} \cdot cos\phi_{2} \cdot \sin^2(\Delta \lambda/2)
    \label{haversine1}
\end{equation}

\begin{equation}
    c = 2 \cdot a\tan 2(\sqrt{a}, \sqrt{1-a})
    \label{haversine2}
\end{equation}

\begin{equation}
    d = R \cdot c
    \label{haversine3}
\end{equation}

where $φ$ is latitude, $λ$ is longitude, R is earth’s radius (mean radius = 6,371km), and \textit{d} represents the distance between two spatial attributes. 

Textual attributes are usually qualitative in nature. However, in some cases, when there are very limited textual attributes are available with an image, LSA fails to derive context from those attributes. In that cases, we use Jaccard similarity measure to find the difference in textual attributes: 

\begin{equation}
    \Delta c = \frac{|c_o \cap c_m|}{|c_o \cup c_m|}
    \label{eq13}
\end{equation}

\noindent where \textit{c\textsubscript{o}} is the original text and \textit{c\textsubscript{m}} denotes modified text.

Quantitative changes in spatio-temporal and contextual attributes may have different scales. Therefore, we first normalize the differences to 0-1 range using the following equations:

\begin{equation}
\Delta \hat{\zeta} = \frac{\Delta \zeta _i - min(\Delta \zeta, \Delta \tau, \Delta c)}{max(\Delta \zeta, \Delta \tau, \Delta c) - min(\Delta \zeta, \Delta \tau, \Delta c)}
\end{equation}
where $\hat{\zeta}$ represents the normalized value of spatial distance.
$Δτ$ and $Δc$ can be normalized similarly.

\subsection{Intention of Modifications}
Every modification in an image may not be a fake. \textit{We define `fake' as modifications that make an image harmful to be used in a specific application}. 
Fakeness is highly context-sensitive. An image which is fake in one context may not be fake for another context. For instance, a fake road accident image may not be possibly fake for other contexts. Therefore, an extensive investigation of the identified image service modifications is required to determine the level of fakeness in a modified image. 
In this respect, we define \textit{intention} as a key parameter to determine the likelihood of an image to be fake.
\textit{Intention refers to the purpose of introducing modifications in an image service}. Intentions can be ill or well. 
Therefore, modifications may either be \textit{ill-intentioned} or \textit{well-intentioned}~\cite{lowe2018self}. For instance, improving the outlook of a picture are usually \textit{well-intentioned} changes. \textit{Well-intentioned} changes do not usually need to be hidden because they are a positive addition to the image. Whereas, \textit{ill-intentioned} modifications are normally concealed from the viewers to hide their evidences. Although, \textit{well-intentioned} changes are not supposed to be harmful but they may change the context of a reconstructed scene. Therefore, all modifications in crowdsourced \textit{image services} should be evaluated before reconstructing a scene.
We propose a multi-layer analysis on image service modifications to reach to a conclusion about image's legitimacy and fakeness as shown in Figure~\ref{fig:framework}. The proposed framework considers semantic closeness of the identified modifications and \% obfuscation (attempt to hide the facts) to estimate the intention of changes.

We consider intention as a spectrum to provide an estimate about the likelihood of a modification to be \textit{ill-intentioned} or \textit{well-intentioned}. We also propose a method to translate intention of modifications into the degree of fakeness. 
A change which turns out to be a fake for one scene, may not be fake for another scene. We propose to estimate \textit{intention} based on the change in the semantics of an image service, and \% obfuscation. 
We consider all the possible combinations of extrinsic modifications ($Δζ$, $Δτ$ and $Δc$) in an image service to derive intention as shown in Table~\ref{Intention}.

\begin{table}[htbp]
\caption{Deriving the Intention}
\begin{center}
\begin{tabular}{|c|c|c|c|c|}
\hline
\textbf{$Δζ$} & \textbf{$Δτ$} & \textbf{$Δc$} & \textbf{Semantic Change} &\textbf{Intention} \\ \hline

$\downarrow$ & $\downarrow$ & $\downarrow$ & $\downarrow$ & + \\
$\downarrow$ & $\downarrow$ & $\uparrow$ & $\uparrow$ & - \\
$\downarrow$ & $\uparrow$ & $\downarrow$ & $\downarrow$ & + \\
$\downarrow$ & $\uparrow$ & $\downarrow$ & $\uparrow$ & - \\
.&. & . & . & . \\
.&. & . & . & . \\
\hline
 
\end{tabular}
\label{Intention}
\end{center}
\begin{tablenotes}
        \item[1] $\downarrow$ refers to a lower value. 
        \item[2] $\uparrow$ refers to a higher value. 
        \item[3] + refers to inclination towards well-intention.
        \item[4] - refers to inclination towards ill-intention.
    \end{tablenotes}
\end{table}

Semantic change in Table~\ref{Intention} represents the overall change in the semantics of an image.
Intention of a few samples is first labeled using fact checked images. A machine learning algorithm is then trained on these samples which can be later used to derive the expression for \textit{intention}. Intention can be expressed as a function of modifications in an image.

\begin{equation}
Intention = \Theta (\Delta \zeta, \Delta \tau, \Delta c)
\label{intentEq}
\end{equation}

Semantic change is implied in the expression of intention by giving bias to the changes having higher impact. Weights in the expression of $Θ$ are computed based on the significance of $Δζ$, $Δτ$ and $Δc$. 
Precision of attributes is also taken into consideration while assigning the weights. For instance, \textit{GPS coordinates} have high precision as compared to \textit{name of a country}. Therefore, GPS coordinates are assigned higher weight. Table~\ref{Rules} shows rules to assign weights. 

\begin{table}[htbp]
\caption{Rules for Assigning Weights}
\begin{center}
\begin{tabular}{|c|c|c|c|}
\hline
& \textbf{Significance} & \textbf{Precision} & \textbf{Weights} \\ \hline

\parbox[t]{2mm}{\multirow{7}{*}{\rotatebox[origin=c]{90}{\textbf{Attributes}}}} 
& High & $>$90\% & 5 \\
& High & 80-90\% & 4 \\
& Medium & 60-70\% & 3 \\
& Low & $<$90\% & 2 \\
& ...... & .. & . \\
& ...... & .. & . \\
\hline
 
\end{tabular}
\label{Rules}
\end{center}
\begin{tablenotes}
        \item[1] Significance is assumed to have 3 levels: High, medium and low. 
        \item[2] $Δζ$, $Δτ$ and $Δc$ are assigned weights ranging  1 to 5.
        \item[3] A changing threshold concept can be used to define thresholds \cite{umair2023energy, umair2020energy}.
    \end{tablenotes}
\end{table}

\begin{figure}[htbp]
\centerline{\includegraphics[width = 1 \columnwidth]{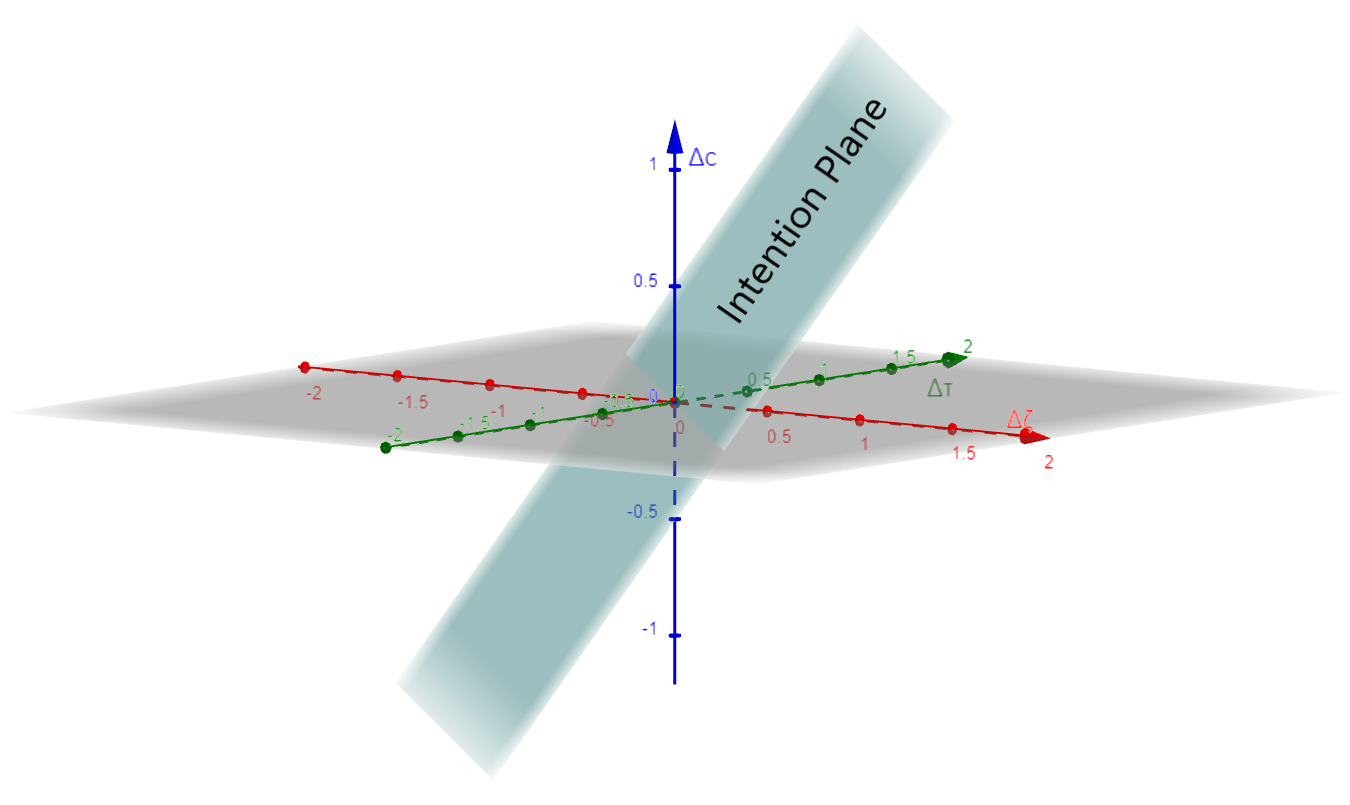}}\caption{A Planar Region representing Intensity of Intention}
\label{IntPlane}
\end{figure}

Different levels of intention can be defined e.g., extremely ill, ill and well intention etc. Equation~\ref{intentEq} represents a plane in 3d for each value of intention as shown in Figure~\ref{IntPlane}. Different planes represent different intensities of intention. Modifications lying close to ill-region represent the likelihood of being an ill-intention change and vice versa.   
Integral can be applied on the equation of each plane to convert it into a region. It helps in labeling each set of modifications with a value of intention. The following expression computes the area under the plane.

\begin{equation}
Intention Area = \int \int_{}^{}\Theta d\Delta \zeta d\tau
\end{equation}



\subsubsection{Finding the Nearest Intention Plane}
Intention planes represent the ground truth about intention. We leverage these planes to estimate the intention of underlying modifications. In this respect, we find the nearest plane to each set of modifications. We compute the shortest distance of all modified images from each plane. The plane having the minimum shortest distance is assigned to the corresponding set of modifications. The following equation is used to find the shortest distance:

\begin{equation}
d = \frac{|a\Delta \zeta + b\Delta \tau + c\Delta c + d|}{\sqrt{a^2 + b^2 + c^2}}
\label{shortestdistance}    
\end{equation}

\noindent where \textit{a, b} and \textit{c} represents the coefficients of the intention.

\subsubsection{Clustering}
Modifications can be clustered before finding nearest intention plane to a modified image. The nearest intention plane is then found for the whole cluster which reduces the number of computations required to estimate intention. We use KMeans clustering to cluster the closely lying modifications. 

\subsubsection{Proposed Algorithm}\label{Algorithm}
The following algorithm provides a pseudo code of the proposed algorithm. 

\begin{algorithm}
\caption{: Algorithm to Estimate Intention}
\small
\begin{algorithmic}[1]
\State \textbf{Input}: $ζ$, $τ$, $c$ 
\State \textbf{Output}: Intention (\textit{I})
\State Compute $Δζ$, $Δτ$ and $Δc$
\State Normalize $Δζ$, $Δτ$ and $Δc$
\State Apply binomial regression to predict \textit{intention (I)} based on modifications and the overall semantic change
\State Use \textit{I} to find equation of different intention planes
\State Perform clustering
\State Find shortest distance \textit{d} of each cluster from every plane
\State return \textit{I} with minimum distance
\end{algorithmic}
\end{algorithm}

\subsubsection{Translating Intention into Fakeness}
\begin{figure*}[htbp]
\centerline{\includegraphics[width = 0.7\textwidth]{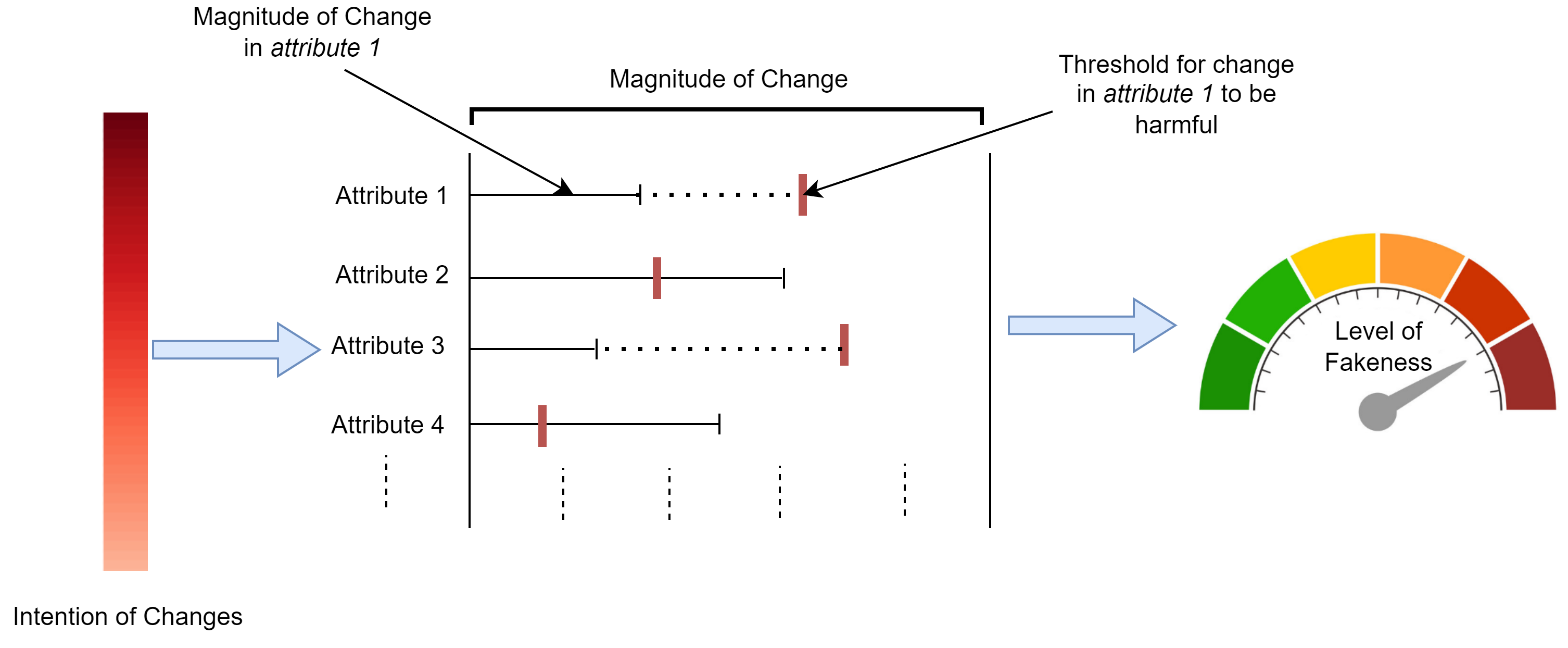}}\caption{Translating intention into fakeness}
\label{fig:Int_to_fake}
\end{figure*}

The proposed framework provides a detailed analysis on the intention of modifications to ascertain fake image services. Afterwards, an expert's opinion can be leveraged to conclude about the fakeness of a picture. Alternatively, we can train a machine learning algorithm to translate intention in to fakeness based on the previous examples. The key difference between the intention and fakeness is the context. Intention is based on a more generic context whereas fakeness is determined in a very specific context. We provide intensity of modifications in each attribute to the experts or a machine learning algorithm. Experts provide their opinion that which of the changes may be harmful in a given context along with a confidence score. Figure~\ref{fig:Int_to_fake} shows the supposed analysis provided by the experts. 

Contextual attributes usually have a wide range of keywords. Repeating the aforementioned analysis for each word can be cumbersome. Therefore, we employ TF-IDF (Term Frequency - Inverse Document Frequency) to determine the significance of modified textual keywords. TF-IDF provides an idea about the relevancy of the keywords in a specific context. The more relevant a keyword to a particular context, the more significant that attribute is.

\section{Experiments and Results}
This section explains data acquisition, implementation and results of the proposed methodology.
\subsection{Dataset}
\subsubsection{Context-Based Dataset}
We evaluate the proposed framework on a real dataset consisting of 1000 images. The images are collected from multiple datasets used in~\cite{boididou2018detection}. We also collect images from different social media platforms (e.g., Twitter, Facebook and Instagram etc.) and copyright free image repositories (e.g., Shutterstock, Unsplash and Pexels etc.). The images are collected for 5 different application domains i.e., road accidents, crime scenes, violent scenes, natural disasters and public gatherings. We manipulate the metadata of these images to introduce different types of systematic changes in the images. We only consider extrinsic non-functional attributes to create these changes. Hence, the modifications introduced in the images are only extrinsic in nature as per the scope of this paper. However, there can be a few intrinsic changes in the images which may be reflected in the extrinsic non-functional attributes. The changes introduced in the metadata are systematic in nature. It is ensured that the changes serve a specific purpose. In this respect, different combinations of changes in spatial, temporal and contextual attributes are introduced. It is ensured that the changes introduced are both well-intentioned and ill-intentioned. The original images with metadata serve as the ground truth in the evaluation process.

\subsubsection{Image Verification Corpus}

We employ another real image metadata dataset for our experiments~\cite{drewnoakes2020}. This dataset comprises a diverse collection of images and videos, each accompanied by its respective metadata. This dataset offers a comprehensive representation encompassing a wide array of subjects, scenes, and visual attributes. To extract metadata, we utilize the metadata-extractor library, a Java-based tool designed for reading metadata from image files. The information extracted includes details like \textit{camera make} and \textit{model}, \textit{image dimensions}, \textit{capture date} and \textit{time}, \textit{GPS coordinates}, and other technical specifications. It's important to note that while the dataset contains images, our focus is solely on their associated metadata. We generate various versions of images based on this metadata dataset leveraging ChatGPT's proficiency in natural language generation and contextual comprehension. Furthermore, we employ ChatGPT to systematically alter the sample metadata of these distinct image versions. Clear instructions are provided to ChatGPT to induce specific variations in the metadata of different versions of an image. For example, we introduce inconsistencies in attributes like \textit{shutter speed} and \textit{exposure time} in some images. Due to its proficient language generation capabilities, ChatGPT can produce modified metadata, including updated timestamps, adjusted camera settings, or edited descriptions, signifying that the image has undergone alterations. As ChatGPT is trained on real world wide web corpus, the modifications introduced by ChatGPT cannot be regarded completely as synthetic changes.

\begin{figure}[htbp]
\centerline{\includegraphics[width = 0.7\columnwidth]{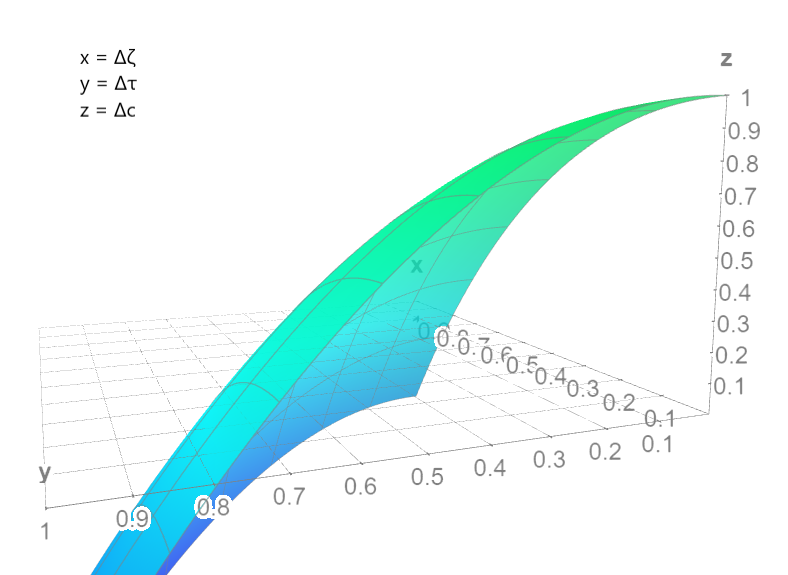}}\caption{Ground Truth of Intention}
\label{trueIntPlane}
\end{figure}
\subsection{Implementation}

We create a python script to perform LSA on the non-functional attributes of an image service. 

\subsubsection{Intention of Modifications}

We propose a machine-learning-based solution to estimate intention from previous samples. We adopt the federated learning approach proposed in \cite{akram2023chained} to train the model. However, we do not always have the labeled data. In that case, we can take an expert opinion to label a few training samples.
We leverage state-of-the-art literature to estimate intention for a few samples. We label different intention score to modified images based on the different combinations of changes in spatio-temporal and contextual attributes. We then apply multinomial regression to predict the intention score for the rest of the dataset. These intention scores represent different intensities of intention. We define 5 different levels of intention: extremely-ill, very-ill, moderately-ill, border-line and well-intention. These values form a plane in a 3d Cartesian space for each value of intention. These planes serve as the ground truth for intention. For the sake of simplicity, we only show one plane in Figure~\ref{trueIntPlane}. It shows a plane with the highest intensity of ill-intention i.e., extremely ill.  


\subsubsection{Clustering}
\begin{figure}[htbp]
\centerline{\includegraphics[width =.8 \columnwidth]{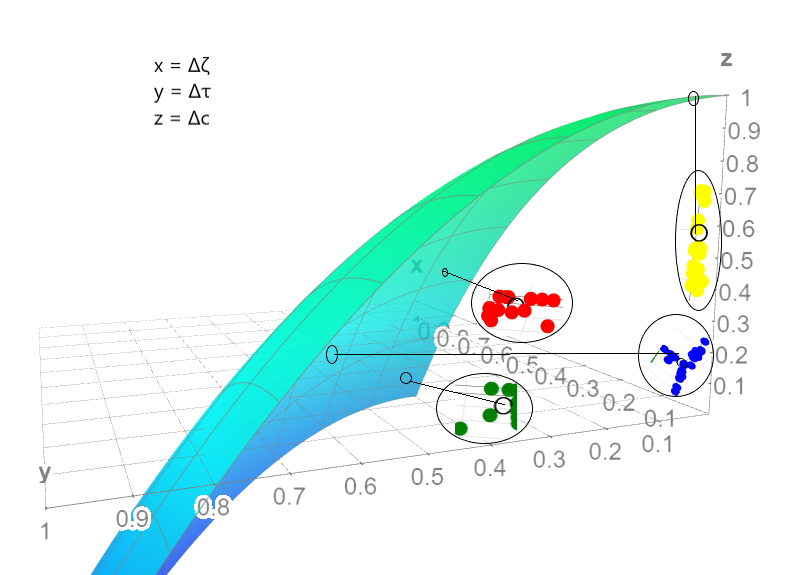}}\caption{Shortest Distance of Clusters from Intention Plane}
\label{clusters-plane}
\end{figure}

Once we have find out the different intensities of intention in the Cartesian space, the next step is to plot the data in the same Cartesian space. As stated earlier, there can be multiple intention regions. The points lying close to the ill regions will more likely be ill-intentioned changes and vice versa. We first perform KMeans clustering (as used in \cite{rizvi2021clustering}) on the combination of modifications. The intention can then be computed for each cluster rather than for each individual sample. It reduces the run time complexity of the algorithm. Figure~\ref{clusters-plane} shows the clusters formed for the modified images in the context of road accidents.


\subsubsection{Finding the Nearest Intention Plane}
Once the modifications have been clustered, the next step is to find the nearest intention plane to the clusters. We need to find the distance of clusters from each plane to find the nearest plane. We use equation~\ref{shortestdistance} to find the distance of clusters from each plane.
Figure~\ref{clusters-plane} shows the distance of clusters from the plane.

\subsubsection{Translating Intention in to Fakeness}
Fakeness is context-sensitive as compared to intention which is context-free. A change which is ill-intention may not be fake in a specific context. Intention refers to significance of the combination of overall changes i.e., $\Delta \zeta$, $\Delta \tau$ and $\Delta c$. Whereas, fakeness refers to the severity of changes in a specific context. An insignificant change in $\Delta \tau$ and $\Delta c$ may still be harmful in a specific context. Therefore, an intense analysis from the domain expert is required to comment on the fakeness of the picture. We propose a machine-learning-based solution to translate intention into fakeness depending upon the changes in individual attributes along with their intention. In that respect, we label fakeness score for few samples which serves as the training set. We train a machine learning model using a python script. It return the fakeness score for each set of non-functional attributes i.e., $\Delta \tau$ and $\Delta c$.
\subsection{Results}
We test our methodology on the three metrics: accuracy, precision and run-time efficiency.

\subsubsection{Accuracy}
\begin{figure}[htbp]
\centerline{\includegraphics[width = 1\columnwidth]{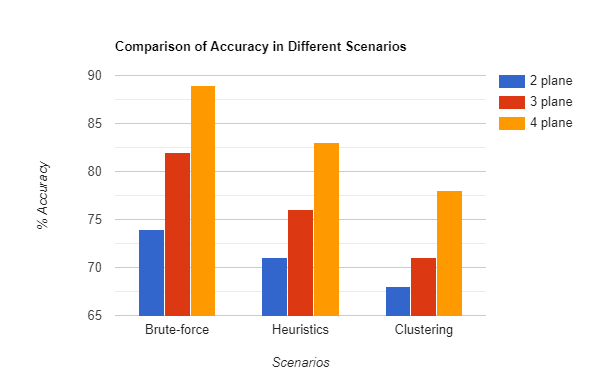}}\caption{Accuracy of the Proposed Approach on Context-Based Dataset}
\label{Accuracy}
\end{figure}

\begin{figure}[htbp]
\centerline{\includegraphics[width = 0.75\columnwidth]{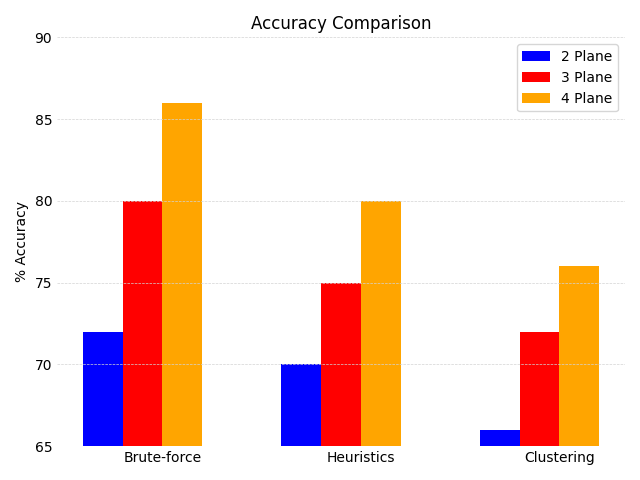}}\caption{Accuracy of the Proposed Approach on Image Verification Corpus}
\label{AccuracyImageVerificationCorpus}
\end{figure}

As mentioned earlier, there will be multiple intention planes in the 3d Cartesian space representing different intensities of intention. The accuracy is computed as the percentage of samples closest to the correct intention plane they belong to. We test the accuracy of the proposed methodology for 2, 3 and 4 planes. We can choose any number of intention planes while training the model. We test our proposed clustering with bruteforce and basic heuristics. Heuristics-based approach don't perform clustering but try to minimize the number of iterations required to find the nearest plane. Figure~\ref{Accuracy} shows the percentage accuracy of these approaches in different scenarios tested on the context-based dataset. Bruteforce approach shows 89\% accuracy with 4 planes. Whereas, heuristics and clustering show 83\% and 78\% accuracy respectively. We also test the accuracy of the proposed approach on image verification corpus (refer to Figure~\ref{AccuracyImageVerificationCorpus}). The accuracy drops slightly on this dataset. The reason behind this drop in accuracy could be the unavailability of substantial metadata tags with each image. Overall, the results indicate that the brute force has the best accuracy. The drop in accuracy may be attributed to the limited availability of substantial metadata tags associated with each image. Overall, The accuracy increases with the increase in the number of intention planes.

\begin{figure}[htbp]
\centerline{\includegraphics[width = 0.75\columnwidth]{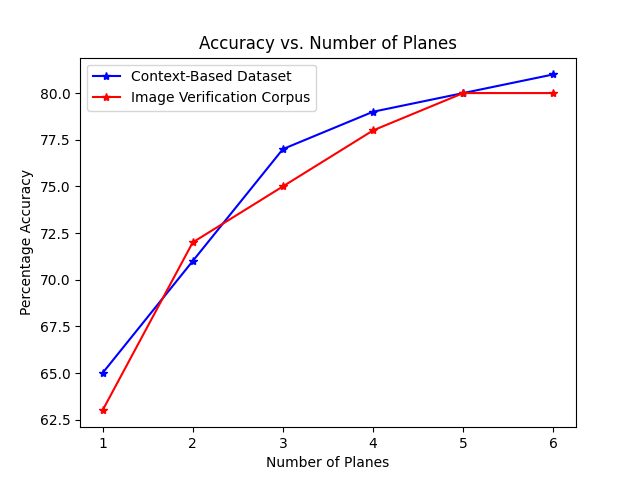}}\caption{Accuracy of the Proposed Approach}
\label{AccuracyTrend}
\end{figure}

Figure~\ref{AccuracyTrend} shows the increase in accuracy with the increase in intention planes for both datasets. It can be noted that the rate of increase in accuracy decreases after increasing the vector size up to a certain extent.

\subsubsection{Precision}
\begin{figure}[htbp]
\centerline{\includegraphics[width = 0.7\columnwidth]{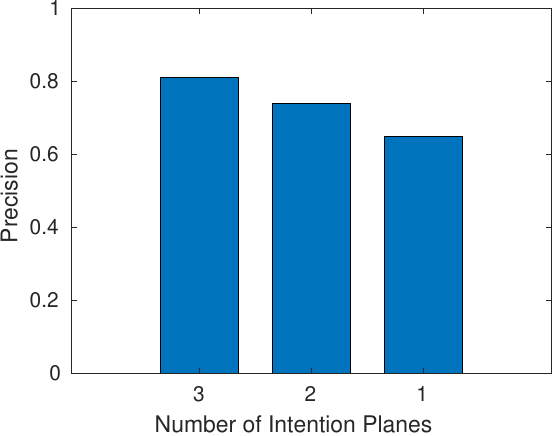}}\caption{Precision of Detecting Ill-intentioned Changes}
\label{Precision}
\end{figure}

We use precision to report the correctly identified ill-intentioned changes. Figure~\ref{Precision} shows the precision of the proposed approach for different number of intention planes. Precision increases with the increase in number of planes.

\subsubsection{Run-Time Efficiency}
We report the run-time efficiency of the proposed approach in terms of time consumed by the algorithm and run-time complexity to train the algorithms. Run-time is reported in nanoseconds (a SI unit of time equal to $10^{-9}$ seconds). Table~\ref{timeEff} shows the run-times of different approaches for both datasets. The framework takes slightly more time for image verification corpus because of data cleaning and preprocessing. 
Clustering is the most efficient approach in terms of run-time efficiency. Whereas, the overhead of training in clustering is O(N\textsuperscript{2}). Heuristics-based approach has a better run-time efficiency as compared to baseline whereas, its run-time complexity is O(N).

\begin{table}[htbp]
\caption{Run Time Efficiency}
\begin{center}
\begin{tabular}{|p{3cm}|c|c|c|}
\hline
\textbf{Time consumed (ns)}&\textbf{Brute-force} & \textbf{Heuristics} & \textbf{Clustering} \\ \hline

\textbf{Context-Based Dataset} & 28000 & 14000 & 1010 \\
\hline
\textbf{Image Verification Corpus} & 31000 & 15400 & 1125 \\
\hline
\textbf{Run-time complexity} & 1 & O(N) & O($N^2$) \\
\hline
\end{tabular}
\label{timeEff}
\end{center}
\end{table}

\vspace{-1em}
\subsubsection{Comparison with State-of-the-Art}

\begin{figure}[htbp]
\centerline{\includegraphics[width = 0.7\columnwidth]{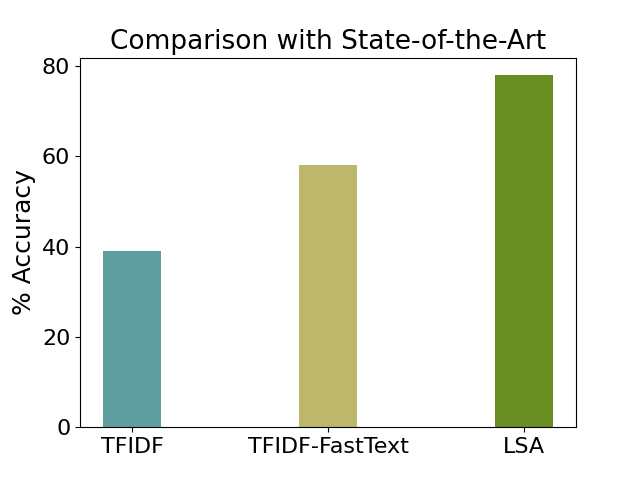}}\caption{Comparison with the state-of-the-art}
\label{SOTA}
\end{figure}

We compare the performance of the proposed approach with a state-of-the-art. As stated earlier, our proposed approach is based on Latent Semantic Analysis (LSA) to derive semantics of the non-functional attributes of an image service. We compare the performance of our LSA-based framework with a recently proposed approach named as TFIDF-FastText~\cite{chawla2023text}. TFIDF-FastText classifies semantically similar short texts. It uses a fusion word-embedding technique – TF-IDF weighted average FastText, to explore the lexical and semantic features of the text. It results in deriving quantitative semantic representation of word tokens. TF-IDF can capture the most descriptive words in a sentence which helps in the efficient clustering of text into classes. The FastText model facilitates the representation of text in a continuous, low-dimensional space, and can record the semantics of a text. Figure~\ref{SOTA} provides a comparison of the accuracy achieved by the proposed approach and the state-of-the-art. TF-IDF achieves 39\% accuracy. However, fusing TFIDF with FastText increases the accuracy up to  58\%. Whereas, the proposed LSA-based approach outperforms TFIDF-FastText achieving an accuracy of 78\%. LSA is more effective as it identifies the latent semantic structure in an image metadata. Whereas, TF-IDF is based on the frequency of words.

\vspace{-1em}

\section{Conclusion}
A novel framework is proposed in this paper to ascertain fake crowdsourced images based on the changes in image metadata. An image is abstracted as a service. The proposed approach is unique in the sense that it relies completely on the changes in image's non-functional attributes, i.e., image metadata, to determine the trustworthiness of crowdsourced image services. Intention is defined as an important parameter to reflect the likelihood of an image being fake. A novel framework is proposed to estimate the intention of underlying modifications. A clustering approach is then used to determine the likelihood of an image being a fake. The results show high accuracy and run-time efficiency.

\vspace{-1em}
\section*{Acknowledgment}

This research was partly made possible by LE220100078 and DP220101823 grants from the Australian Research Council. The statements made herein are solely the responsibility of the authors.

\vspace{-1em}

\bibliographystyle{IEEEtran}
\bibliography{bibliography.bib}

\begin{thebibliography}{10}
\providecommand{\url}[1]{#1}
\csname url@samestyle\endcsname
\providecommand{\newblock}{\relax}
\providecommand{\bibinfo}[2]{#2}
\providecommand{\BIBentrySTDinterwordspacing}{\spaceskip=0pt\relax}
\providecommand{\BIBentryALTinterwordstretchfactor}{4}
\providecommand{\BIBentryALTinterwordspacing}{\spaceskip=\fontdimen2\font plus
\BIBentryALTinterwordstretchfactor\fontdimen3\font minus \fontdimen4\font\relax}
\providecommand{\BIBforeignlanguage}[2]{{%
\expandafter\ifx\csname l@#1\endcsname\relax
\typeout{** WARNING: IEEEtran.bst: No hyphenation pattern has been}%
\typeout{** loaded for the language `#1'. Using the pattern for}%
\typeout{** the default language instead.}%
\else
\language=\csname l@#1\endcsname
\fi
#2}}
\providecommand{\BIBdecl}{\relax}
\BIBdecl

\bibitem{griffis2014use}
H.~M. Griffis, A.~S. Kilaru, R.~M. Werner, D.~A. Asch, J.~C. Hershey, S.~Hill, Y.~P. Ha, A.~Sellers, K.~Mahoney, and R.~M. Merchant, ``Use of social media across us hospitals: descriptive analysis of adoption and utilization,'' \emph{Journal of medical internet research}, vol.~16, no.~11, p. e264, 2014.

\bibitem{liu2011using}
X.~Liu, R.~Troncy, and B.~Huet, ``Using social media to identify events,'' in \emph{Proceedings of the 3rd ACM SIGMM international workshop on Social media}, 2011, pp. 3--8.

\bibitem{rosi2011social}
A.~Rosi, M.~Mamei, F.~Zambonelli, S.~Dobson, G.~Stevenson, and J.~Ye, ``Social sensors and pervasive services: Approaches and perspectives,'' in \emph{2011 IEEE international conference on pervasive computing and communications workshops (PERCOM Workshops)}.\hskip 1em plus 0.5em minus 0.4em\relax IEEE, 2011, pp. 525--530.

\bibitem{gu2016twitter}
Y.~Gu, Z.~S. Qian, and F.~Chen, ``From twitter to detector: Real-time traffic incident detection using social media data,'' \emph{Transportation research part C: emerging technologies}, vol.~67, pp. 321--342, 2016.

\bibitem{aamir2018social}
T.~Aamir, H.~Dong, and A.~Bouguettaya, ``Social-sensor composition for scene analysis,'' in \emph{International Conference on Service-Oriented Computing}.\hskip 1em plus 0.5em minus 0.4em\relax Springer, 2018, pp. 352--362.

\bibitem{shen2019fake}
C.~Shen, M.~Kasra, W.~Pan, G.~A. Bassett, Y.~Malloch, and J.~F. O’Brien, ``Fake images: The effects of source, intermediary, and digital media literacy on contextual assessment of image credibility online,'' \emph{New media \& society}, vol.~21, no.~2, pp. 438--463, 2019.

\bibitem{gupta2013faking}
A.~Gupta, H.~Lamba, P.~Kumaraguru, and A.~Joshi, ``Faking sandy: characterizing and identifying fake images on twitter during hurricane sandy,'' in \emph{Proceedings of the 22nd international conference on World Wide Web}, 2013, pp. 729--736.

\bibitem{aamir2018stance}
T.~Aamir, H.~Dong, and A.~Bouguettaya, ``Stance and credibility based trust in social-sensor cloud services,'' in \emph{International Conference on Web Information Systems Engineering}.\hskip 1em plus 0.5em minus 0.4em\relax Springer, 2018, pp. 178--189.

\bibitem{aamir2018trust}
T.~Aamir~et al, ``Trust in social-sensor cloud service,'' in \emph{2018 IEEE International Conference on Web Services (ICWS)}.\hskip 1em plus 0.5em minus 0.4em\relax IEEE, 2018, pp. 359--362.

\bibitem{colliander2019fake}
J.~Colliander, ``“this is fake news”: Investigating the role of conformity to other users’ views when commenting on and spreading disinformation in social media,'' \emph{Computers in Human Behavior}, vol.~97, pp. 202--215, 2019.

\bibitem{ghosh2017detection}
P.~Ghosh, V.~Morariu, B.-C.~I. Larry~Davis \emph{et~al.}, ``Detection of metadata tampering through discrepancy between image content and metadata using multi-task deep learning,'' in \emph{Proceedings of the IEEE Conference on Computer Vision and Pattern Recognition Workshops}, 2017, pp. 60--68.

\bibitem{riggs2018image}
C.~Riggs, T.~Douglas, and K.~Gagneja, ``Image mapping through metadata,'' in \emph{2018 Third International Conference on Security of Smart Cities, Industrial Control System and Communications (SSIC)}.\hskip 1em plus 0.5em minus 0.4em\relax IEEE, 2018, pp. 1--8.

\bibitem{aamir2020social}
T.~Aamir, H.~Dong, and A.~Bouguettaya, ``Social-sensor composition for tapestry scenes,'' \emph{IEEE Transactions on Services Computing}, 2020.

\bibitem{chae2012spatiotemporal}
J.~Chae, D.~Thom, H.~Bosch, Y.~Jang, R.~Maciejewski, D.~S. Ebert, and T.~Ertl, ``Spatiotemporal social media analytics for abnormal event detection and examination using seasonal-trend decomposition,'' in \emph{2012 IEEE Conference on Visual Analytics Science and Technology (VAST)}.\hskip 1em plus 0.5em minus 0.4em\relax IEEE, 2012, pp. 143--152.

\bibitem{slabaugh2001survey}
G.~Slabaugh, R.~Schafer, T.~Malzbender, and B.~Culbertson, ``A survey of methods for volumetric scene reconstruction from photographs,'' in \emph{Volume Graphics 2001}.\hskip 1em plus 0.5em minus 0.4em\relax Springer, 2001, pp. 81--100.

\bibitem{aamir2017social2}
T.~Aamir, A.~Bouguettaya, H.~Dong, S.~Mistry, and A.~Erradi, ``Social-sensor cloud service for scene reconstruction,'' in \emph{International Conference on Service-Oriented Computing}.\hskip 1em plus 0.5em minus 0.4em\relax Springer, 2017, pp. 37--52.

\bibitem{patel2022fake}
M.~Patel, J.~Padiya, and M.~Singh, ``Fake news detection using machine learning and natural language processing,'' in \emph{Combating Fake News with Computational Intelligence Techniques}.\hskip 1em plus 0.5em minus 0.4em\relax Springer, 2022, pp. 127--148.

\bibitem{singh2021predicting}
B.~Singh and D.~K. Sharma, ``Predicting image credibility in fake news over social media using multi-modal approach,'' \emph{Neural Computing and Applications}, pp. 1--15, 2021.

\bibitem{hsu2020deep}
C.-C. Hsu, Y.-X. Zhuang, and C.-Y. Lee, ``Deep fake image detection based on pairwise learning,'' \emph{Applied Sciences}, vol.~10, no.~1, p. 370, 2020.

\bibitem{singh2021image}
B.~Singh and D.~K. Sharma, ``Image forgery over social media platforms-a deep learning approach for its detection and localization,'' in \emph{2021 8th International Conference on Computing for Sustainable Global Development (INDIACom)}.\hskip 1em plus 0.5em minus 0.4em\relax IEEE, 2021, pp. 705--709.

\bibitem{yang2021detecting}
J.~Yang, S.~Xiao, A.~Li, G.~Lan, and H.~Wang, ``Detecting fake images by identifying potential texture difference,'' \emph{Future Generation Computer Systems}, vol. 125, pp. 127--135, 2021.

\bibitem{kour2022fast}
V.~Kour, P.~Aggarwal, and R.~Kaur, ``A fast block-based technique to detect copy-move forgery in digital images,'' in \emph{Recent Advances in Artificial Intelligence and Data Engineering}.\hskip 1em plus 0.5em minus 0.4em\relax Springer, 2022, pp. 299--307.

\bibitem{guo2018fake}
Y.~Guo, X.~Cao, W.~Zhang, and R.~Wang, ``Fake colorized image detection,'' \emph{IEEE Transactions on Information Forensics and Security}, vol.~13, no.~8, pp. 1932--1944, 2018.

\bibitem{tanaka2021fake}
M.~Tanaka and H.~Kiya, ``Fake-image detection with robust hashing,'' in \emph{2021 IEEE 3rd Global Conference on Life Sciences and Technologies (LifeTech)}.\hskip 1em plus 0.5em minus 0.4em\relax IEEE, 2021, pp. 40--43.

\bibitem{umair2023detecting}
M.~Umair, A.~Bouguettaya, and A.~Lakhdari, ``Detecting changes in crowdsourced social media images,'' \emph{arXiv preprint arXiv:2310.16848}, 2023.

\bibitem{umair2020multi}
M.~Umair, Z.~Saeed, M.~Ahmad, H.~Amir, B.~Akmal, and N.~Ahmad, ``Multi-class classification of bi-lingual sms using naive bayes algorithm,'' in \emph{2020 IEEE 23rd International Multitopic Conference (INMIC)}.\hskip 1em plus 0.5em minus 0.4em\relax IEEE, 2020, pp. 1--5.

\bibitem{zimmer2019fake}
F.~Zimmer, K.~Scheibe, M.~Stock, and W.~G. Stock, ``Fake news in social media: Bad algorithms or biased users?'' \emph{Journal of Information Science Theory and Practice}, vol.~7, no.~2, pp. 40--53, 2019.

\bibitem{elmaci2021comparative}
M.~Elmaci, A.~N. Toprak, and V.~Aslantas, ``A comparative study on the detection of image forgery of tampered background or foreground,'' in \emph{2021 9th International Symposium on Digital Forensics and Security (ISDFS)}.\hskip 1em plus 0.5em minus 0.4em\relax IEEE, 2021, pp. 1--5.

\bibitem{padilha2021content}
R.~Padilha, T.~Salem, S.~Workman, F.~A. Andal{\'o}, A.~Rocha, and N.~Jacobs, ``Content-based detection of temporal metadata manipulation,'' \emph{arXiv preprint arXiv:2103.04736}, 2021.

\bibitem{simon2022automatic}
G.~Simon, M.~R{\'a}tosi, and G.~Vakulya, ``Automatic measurement of digital cameras’ exposure time using equivalent sampling,'' \emph{IEEE Transactions on Instrumentation and Measurement}, vol.~71, pp. 1--10, 2022.

\bibitem{li2020low}
L.~Li, W.~Li, Q.~Du, and R.~Tao, ``Low-rank and sparse decomposition with mixture of gaussian for hyperspectral anomaly detection,'' \emph{IEEE Transactions on Cybernetics}, vol.~51, no.~9, pp. 4363--4372, 2020.

\bibitem{lowe2018self}
E.~Lowe-Calverley and R.~Grieve, ``Self-ie love: Predictors of image editing intentions on facebook,'' \emph{Telematics and Informatics}, vol.~35, no.~1, pp. 186--194, 2018.

\bibitem{umair2023energy}
M.~Umair, M.~A. Cheema, B.~Afzal, and G.~Shah, ``Energy management of smart homes over fog-based iot architecture,'' \emph{Sustainable Computing: Informatics and Systems}, vol.~39, p. 100898, 2023.

\bibitem{umair2020energy}
M.~Umair and G.~A. Shah, ``Energy management of smart homes,'' in \emph{2020 IEEE International Conference on Smart Computing (SMARTCOMP)}.\hskip 1em plus 0.5em minus 0.4em\relax IEEE, 2020, pp. 247--249.

\bibitem{boididou2018detection}
Detection and visualization of misleading content~on Twitter, ``Boididou, christina and papadopoulos, symeon and zampoglou, markos and apostolidis, lazaros and papadopoulou, olga and kompatsiaris, yiannis,'' \emph{International Journal of Multimedia Information Retrieval}, vol.~7, no.~1, pp. 71--86, 2018.

\bibitem{drewnoakes2020}
D.~Noakes, ``{Database of images from various digital cameras},'' \url{https://github.com/drewnoakes/metadata-extractor-images}, 2019.

\bibitem{akram2023chained}
J.~Akram, M.~Umair, R.~H. Jhaveri, M.~N. Riaz, H.~Chi, and S.~Malebary, ``Chained-drones: Blockchain-based privacy-preserving framework for secure and intelligent service provisioning in internet of drone things,'' \emph{Computers and Electrical Engineering}, vol. 110, p. 108772, 2023.

\bibitem{rizvi2021clustering}
S.~A. Rizvi, M.~Umair, and M.~A. Cheema, ``Clustering of countries for covid-19 cases based on disease prevalence, health systems and environmental indicators,'' \emph{Chaos, Solitons \& Fractals}, vol. 151, p. 111240, 2021.

\bibitem{chawla2023text}
S.~Chawla, R.~Kaur, and P.~Aggarwal, ``Text classification framework for short text based on tfidf-fasttext,'' \emph{Multimedia Tools and Applications}, pp. 1--14, 2023.

\end{thebibliography}

\begin{IEEEbiography}[{\includegraphics[width=1in,height=1.25in]{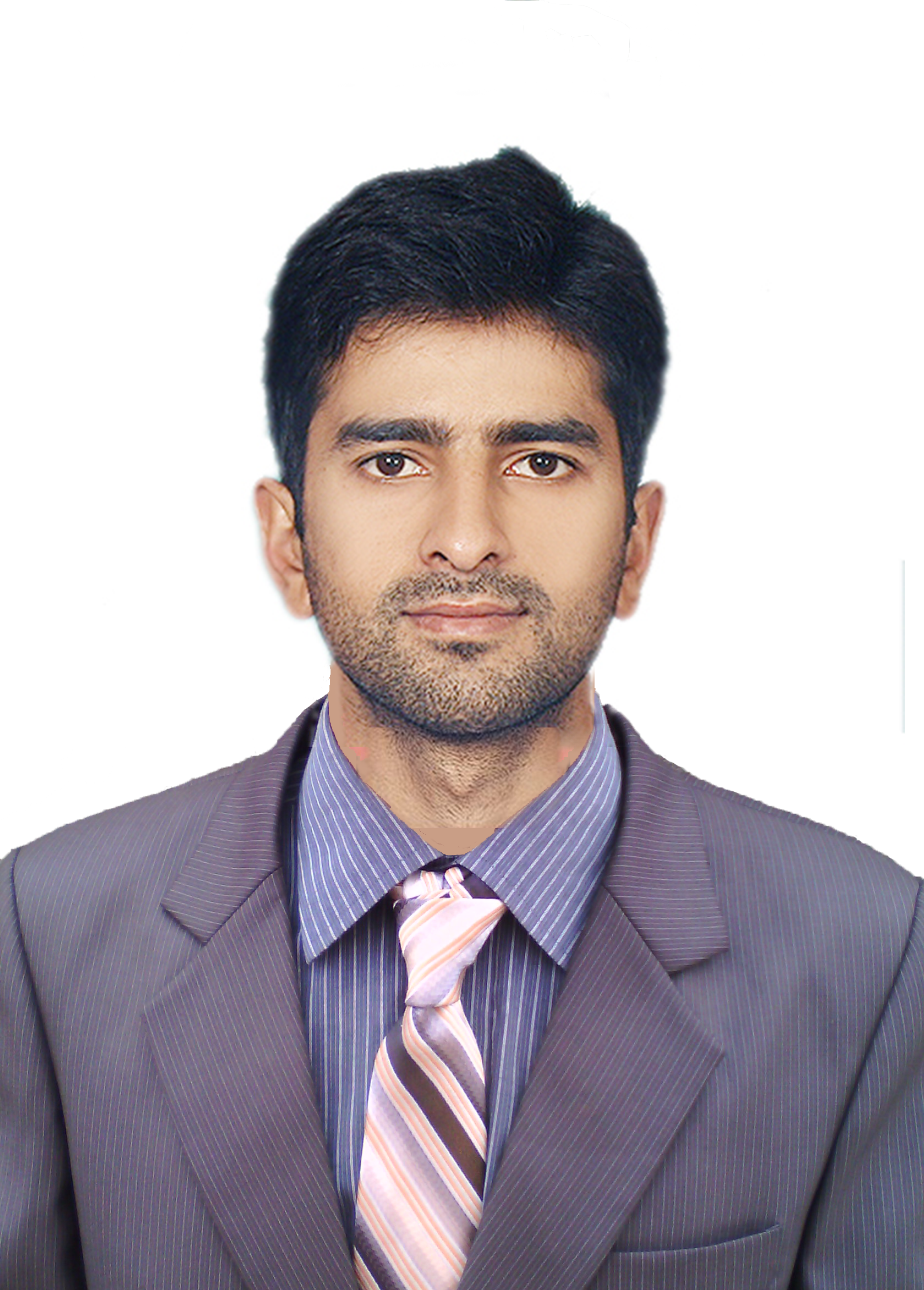}}]{Muhammad Umair}
is a PhD  student at the School of Computer Science, The University of Sydney. He completed his B.Sc. Electrical Engineering and M.Sc. Electrical Engineering from University of Engineering \& Technology (UET) Lahore in 2014 and 2017, respectively. He has worked as a Research Officer at Internet of Things (IoT) lab at Al-Khwarizmi Institute of Computer Sciences, UET Lahore. He has also worked at Sultan Qaboos IT Research lab as a Research Officer. 
\end{IEEEbiography}

\begin{IEEEbiography}[{\includegraphics[width=1in,height=1.25in]{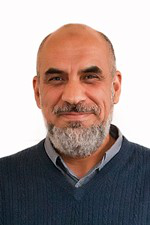}}]{Athman Bouguettaya}
is a Professor in the School of Computer Science at the University of Sydney. He received his Ph.D. in Computer Science from the University of Colorado at Boulder (USA) in 1992. He is or has been on the editorial boards of several journals, including the IEEE Transactions on Services Computing, ACM Transactions on Internet Technology, the International Journal on Next Generation Computing, and VLDB Journal. He is a Fellow of the IEEE and a Distinguished Scientist of the ACM. 
\end{IEEEbiography}

\begin{IEEEbiography}[{\includegraphics[width=1in,height=1.25in]{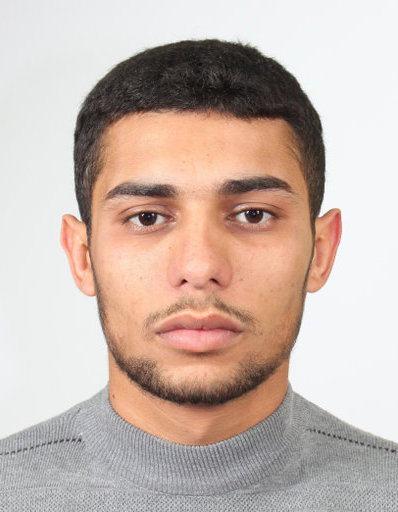}}]{Abdallah Lakhdari}
is a Postdoctoral Fellow in the School of Computer Science at the University of Sydney. He received his Bachelor's degree (2010) and Master's degree (2013) in Computer Science from The University of Laghouat, Algeria. He was a visiting scholar a New Mexico Tech. He worked as a lecturer at the Department of Computer Science at The University of Laghouat, Algeria. His research interests include Social Computing, Crowdsourcing, IoT, and Service Computing.
\end{IEEEbiography}

\end{document}